\documentclass[pdflatex,sn-mathphys-num]{sn-jnl}

\usepackage{graphicx}%
\usepackage{multirow}%
\usepackage{amsmath,amssymb,amsfonts}%
\usepackage{amsthm}%
\usepackage{mathrsfs}%
\usepackage[title]{appendix}%
\usepackage{xcolor}%
\usepackage{textcomp}%
\usepackage{manyfoot}%
\usepackage{booktabs}%
\usepackage{algorithm}%
\usepackage{algorithmicx}%
\usepackage{algpseudocode}%
\usepackage{listings}%

\theoremstyle{thmstyleone}%
\theoremstyle{thmstyletwo}%

\theoremstyle{thmstylethree}%
\newcommand{\switch}{off} 
\newcommand{\toggletext}[1]{%
    \ifthenelse{\equal{\switch}{on}}{#1}{\ignorespaces}%
}

\begin{document}

\title[Article Title]{Multitask finetuning and acceleration of chemical pretrained models for small molecule drug property prediction}

\author[1]{\fnm{Matthew} \sur{Adrian}}\email{matthew.adrian@merck.com}
\equalcont{These authors contributed equally to this work.}

\author[1]{\fnm{Yunsie} \sur{Chung}}\email{yunsie.chung@merck.com}
\equalcont{These authors contributed equally to this work.}

\author[2]{\fnm{Kevin} \sur{Boyd}}\email{kboyd@nvidia.com}

\author[2]{\fnm{Saee} \sur{Paliwal}}\email{saeep@nvidia.com}

\author*[2]{\fnm{Srimukh Prasad} \sur{Veccham}}\email{sveccham@nvidia.com}

\author*[1]{\fnm{Alan C.} \sur{Cheng}}\email{alan.cheng@merck.com}

\affil*[1]{\orgdiv{Modeling \& Informatics}, \orgname{Merck \& Co., Inc.}, \orgaddress{\street{213 E. Grand Ave.}, \city{South San Francisco}, \state{California},  \postcode{94080}, \country{USA}}}

\affil[2]{\orgdiv{BioNeMo}, \orgname{NVIDIA}, \orgaddress{\street{2788 San Tomas Expressway}, \city{Santa Clara}, \state{California}, \postcode{95051}, \country{USA}}}

\newcommand{\SV}[1]{{\color{red}{[Srimukh: #1]}}}
\newcommand{\rename}[1]{{\color{purple}{[renaming #1]}}}

\abstract{Chemical pretrained models, sometimes referred to as foundation models, are receiving considerable interest for drug discovery applications. The general chemical knowledge extracted from self-supervised training has the potential to improve predictions for critical drug discovery endpoints, including on-target potency and ADMET properties. Multi-task learning has previously been successfully leveraged to improve predictive models. Here, we show that enabling multitasking in finetuning of chemical pretrained graph neural network models such as Kinetic GROVER Multi-Task (KERMT), an enhanced version of the GROVER model, and Knowledge-guided Pre-training of Graph Transformer (KGPT) significantly improves performance over non-pretrained graph neural network models. Surprisingly, we find that the performance improvement from finetuning KERMT in a multitask manner is most significant at larger data sizes. Additionally, we publish two multitask ADMET data splits to enable more accurate benchmarking of multitask deep learning methods for drug property prediction. Finally, we provide an accelerated implementation of the KERMT model on GitHub, unlocking large-scale pretraining, finetuning, and inference in industrial drug discovery workflows.}

\maketitle

\section{Introduction}\label{sec:intro}

In drug discovery, predictive machine learning (ML) models are increasingly being used to guide multi-property optimization of small molecules and cyclic peptides. Recent deep learning advances have led to a step-change improvement in the predictive accuracy of models, particularly for ADMET (absorption, distribution, metabolism, excretion, and toxicity) properties. \cite{Caceres2020, Feinberg2020, Walter2024, Ferreira2019} By better leveraging prior experimental data, these improved ML algorithms can lead to better triage of molecular designs prior to experimental synthesis and testing, improved properties in clinical candidates, and ultimately, reduction in cost of drug discovery and improved probability of success for clinical molecules.\cite{Feinberg2020, Beckers2023, VandeWaterbeemd2003, Rinaki2003}

Supervised deep learning architectures, especially graph neural networks (GNNs), are increasingly used in real-world drug discovery pipelines for property prediction \cite{Feinberg2020, Walter2024, Chen2025}. GNNs flexibly learn the latent molecular representations of chemical graphs that are then passed through a feed-forward neural network that learns to predict properties.\cite{Yang2019, Feinberg2020, Heid2024} However, these methods often have limited performance in low-data scenarios\cite{Chen2025} such as for on-target potency. In these cases, more classical methods such as Random Forest are typically used, although the feature curation required for these methods is not trivial. \cite{Chen2025}

Recent developments around chemical pretrained models look to address such limitations in current methods.\cite{Ross2022, Li2023, Rong2020} Though the scale of labeled chemical data is typically limited to $10^{1}$ to $10^{6}$ labeled molecules, the entire chemical space is estimated to be on the order of $10^{60}$ molecules\cite{Bohacek1996}. The subset of known and synthesizable molecules that are unlabeled can be leveraged to pre-train large, self-supervised deep learning models to learn general knowledge and patterns in chemistry. These pretrained models can be subsequently finetuned in a supervised manner on smaller labeled datasets and more effectively learn a specific property prediction task. The general knowledge contained within pre-trained models is believed to generate more expressive molecular representations that could potentially improve performance, especially on small datasets\cite{Ross2022, Li2023, Rong2020}.

In drug property prediction, especially with ADMET properties, many tasks are correlated, presenting an opportunity to use multitask learning, an approach that improves performance based on inductive transfer learning of related tasks. \cite{Caruana1997} Though this has been heavily applied to improve molecular property predictions in deep learning architectures trained from scratch \cite{Wenzel2019, Feinberg2020, Walter2024}, this has not been explored as a finetuning method for chemical pretrained models to our knowledge. Here, we show that multitask-based finetuning improves the performance of KERMT, an enhancement of the chemical pretrained GROVER model\cite{Rong2020}, on both public and internal datasets from Merck \& Co., Inc. (Rahway, NJ, USA) relative to its counterpart single-task models. \textbf{K}inetic GROV\textbf{ER} \textbf{M}ulti-\textbf{T}ask (KERMT) is an enhancement of the GROVER model with distributed pretrained implemented using PyTorch Distributed Data Parallel (DDP). KERMT also contains accelerated finetuning and inference implemented using the \texttt{cuik-molmaker} package \cite{cuikmolmaker}. KERMT also contains automated hyperparameter optimization for finetuning for easier hyperparameter search. These enhancements enable efficient large-scale distributed-GPU training and accelerated finetuning and inference on industry-scale datasets. Contrary to current hypotheses, we show that KERMT outperforms other benchmark models more significantly in large data scenarios. We publish two benchmark multitask ADMET dataset splits for better future assessment of multitasking methods.

\section{Results}
\label{sec:result}

\subsection{Chemical foundation model benchmarking}
\label{ssec:benchmark}

\begin{figure}[h!]
\vspace*{-1.0\baselineskip}
\centering
\includegraphics[width=1.0\textwidth]{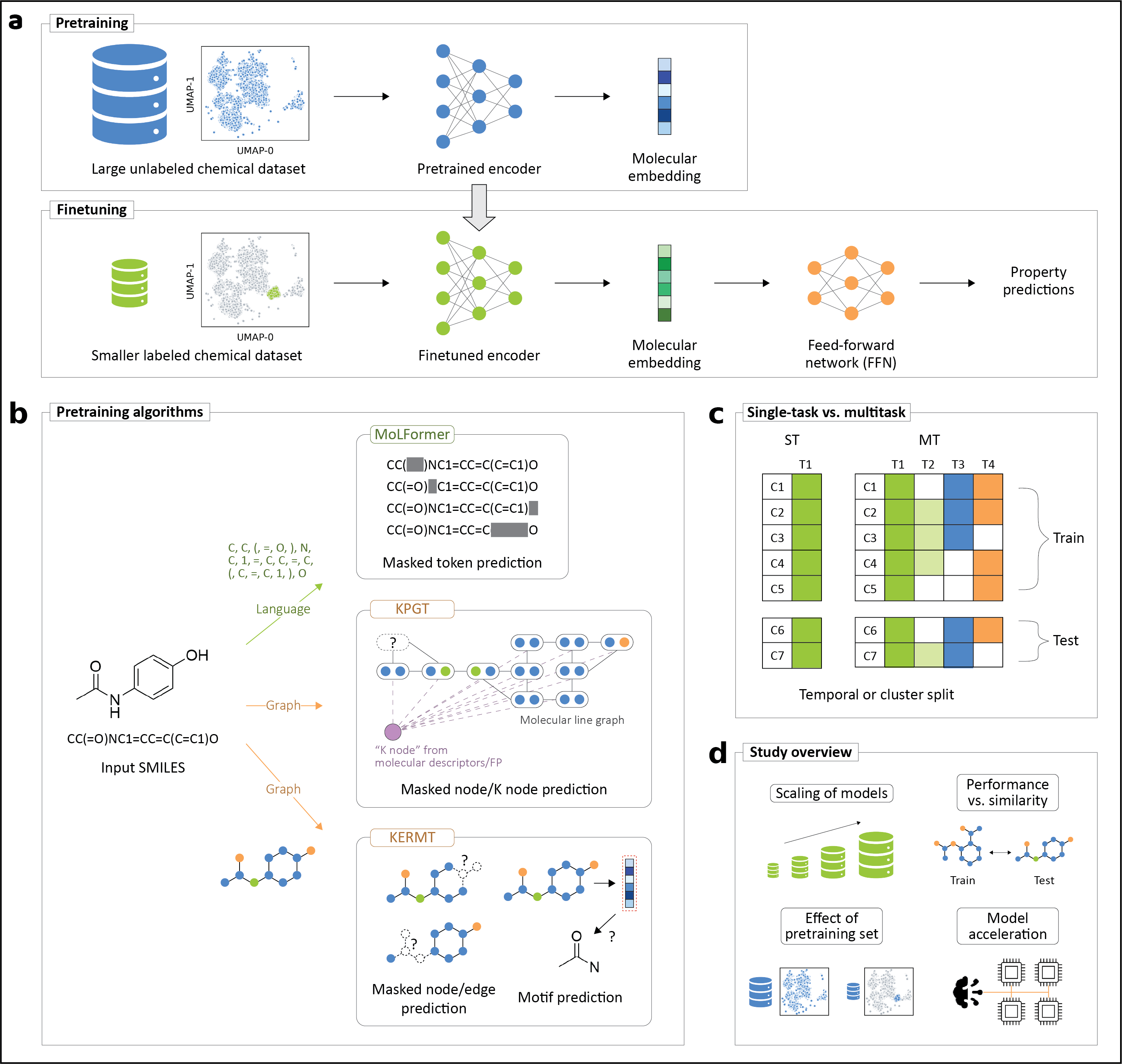}
\caption{\textbf{Schematic overview of the studies.} \textbf{a} Pretraining and finetuning scheme. The parameters of the pretrained chemical encoder are updated during finetuning with labeled data. \textbf{b} Pretraining algorithms investigated. The algorithms are categorized into language-based and graph-based models. \textbf{c} Single-task and multitask scheme. C1 - C7 represent example chemical compounds, and T1 - T4 represent example prediction tasks. For multitask model, the dataset is sparse. The datasets are divided into training and test sets using either temporal or cluster-based splits. \textbf{d} Summary of the investigated studies.}
\label{fig:overview}
\end{figure}

The overview of our study is illustrated in Fig. \ref{fig:overview}. We benchmarked three diverse pretrained encoder models in a single-task (ST) manner first and applied the multitask (MT) finetuning method on the highest performing models. To this end, we evaluated KERMT (a reimplementation of GROVER\cite{Rong2020} with further enhancements), KPGT \cite{Li2023}, and MoLFormer \cite{Ross2022} (Fig. \ref{fig:overview}b); each with a different method of  representing molecules and a different self-supervised training algorithm. KERMT and KPGT both use graph-based transformers while MoLFormer uses a language-based transformer. KERMT represents a molecule as a graph with nodes and edges describing atoms and bonds and is pretrained on 11 million compounds. KPGT uses a molecular-line graph representation with an additional knowledge node, "K-node", to contain molecular descriptors and fingerprints (FP). KPGT is pretrained on 2 million molecules. MoLFormer employs a chemical language transformer with rotational positional embeddings and is pretrained on 100 million compounds. During finetuning, we allowed the weights of both the encoder and feed forward network (FFN) to update for all models, as recommended by the original papers for the pretrained models. For KPGT, we employed two finetuning schemes; the simple finetuning model (referred to as "KPGT") was trained by simply allowing the encoder and FFN to update, while the complex finetuning (referred to as "KPGT\textsubscript{complex}") model utilized the complex finetuning strategies presented in the original paper. These pretrained models are compared against the baseline graph neural network, Chemprop, which is commonly used in real drug discovery settings \cite{Yang2019, Heid2024}. More details on the models can be found in the Methods section.

The models are tested on three ADMET datasets, six single-task on-target potency datasets, and a single multitask kinase on-target potency dataset. The ADMET datasets include the following: (1) internal data from Merck \& Co., Inc. (Rahway, NJ, USA) with 30 ADMET endpoints and 800,733 compounds measured up to 2024, (2) public data with 25 ADMET endpoints and 114,112 compounds collected from the literature \cite{Wenzel2019, Iwata2022, Kim2023, Watanabe2018, Falcon-Cano2022, Esposito2020, Braga2015, Aliagas2022, Perryman2020, Meng2022, Vermeire2022}, (3) public Biogen dataset \cite{Fang2023} with six ADMET endpoints and 3,521 compounds. All ADMET datasets are sparse, with each endpoint containing different amount of data. We chose five to six diverse endpoints from these datasets to benchmark single-task methods and subsequently compared multitask methods on all available endpoints for the best-performing models. \toggletext{The Spearman correlation coefficient heatmap between the assays of each dataset is displayed in Supplementary Fig. 1. The distribution of the maximum Tanimoto similarity score of each test set molecule to the molecules in the training set is presented in Supplementary Fig. 2.}

The sixteen on-target potency datasets consist of internal $\text{EC}_{50}$ and $\text{IC}_{50}$ data for two undisclosed targets from Merck \& Co., Inc. (Rahway, NJ, USA), which we refer to as "Target 1" and "Target 2", as well as public  $\text{IC}_{50}$ datasets from BindingDB\cite{Liu2024} and two literature publications\cite{Li2023,Theisen2024}. The potency datasets have fewer data points compared to the ADMET datasets. The Target 1 and Target 2 datasets each include 744 and 1,163 compounds, the EGFR and BTK datasets each contain 9,462 and 9,337 compounds, and the FGFR1 and HPK1 datasets each contain 12,461 and 4,442 compounds. A multitask kinase dataset contains 26,577 compounds across ten kinase endpoints (ABL1, BRAF, EGFR, ERBB2, KDR, MAPK14, NTRK1, PIK3Ca, PIK3Cg, and RAF1)\cite{Theisen2024}. All internal datasets are split temporally into 80-20 train-test sets, with the test set containing the most recent 20\% compound data and the training set containing all other compounds developed before the 20\% temporal cutoff (around April 2018). The public datasets are split into our own 80-20 train-test cluster split based on clusters generated using compound's PCA-reduced Morgan fingerprints. More details regarding the datasets and splits can be found in the Methods section.

\subsubsection{KERMT multitask outperforms other methods on internal ADMET data}
\label{sssec:internal}

Benchmarking each of the methods on our internal data allows us to observe performance on a larger, more diverse, and more drug-like dataset taken from a single source. Additionally, we are able to evaluate each method on a temporal split; a more difficult task which better simulates the generalization capabilities needed in an industrial drug discovery context by following how the chemical space evolves over time for new drug discovery programs.\cite{Sheridan2013} In temporal split, models are trained on older data for properties to predict the properties of newly developed molecules. We show the benchmarking results in Table \ref{tab:internal}.

\begin{table}[h]
\caption{Internal ADMET data benchmarking of single-task and multitask methods (Pearson $r^2$) on the 80-20 temporal split}\label{tab:internal}
\addtolength{\tabcolsep}{-5pt}
\begin{tabular*}{\textwidth}{@{\extracolsep\fill}lccccc}
\toprule%

 & $\text{P}_\text{app}$ & EPSA & $\text{F}_\text{u,p}$, human & Pgp, rat & MRT, rat \\
Model & $n_{\text{train}}$=41k & $n_{\text{train}}$=7k & $n_{\text{train}}$=18k & $n_{\text{train}}$=20k & $n_{\text{train}}$=53k \\
 \midrule
MolFormer ST & $0.506\pm0.014$ & $0.679\pm0.009$ & $0.534\pm0.032$ & $0.468\pm0.010$ & $\mathbf{0.115\pm0.011}$ \\
Chemprop ST & $0.591\pm0.021$ & $\mathbf{0.816\pm0.007}$ & $\mathbf{0.625\pm0.019}$ & $0.578\pm0.014$ & $0.023\pm0.010$ \\
KPGT ST\textsubscript{complex} & $0.496\pm0.004$ & $0.795\pm0.003$ & $0.574\pm0.015$ & $0.513\pm0.004$ & $0.058\pm0.012$ \\
KPGT ST & $0.596\pm0.004$ & $\mathbf{0.819\pm0.002}$ & $0.598\pm0.012$ & $0.575\pm0.004$ & $0.072\pm0.004$ \\
KERMT ST & $0.641\pm0.012$ & $0.813\pm0.005$ & $0.594\pm0.007$ & $0.605\pm0.006$ & $0.052\pm0.013$ \\
 \midrule
Chemprop MT & $0.657\pm0.014$ & $0.805\pm0.011$ & $\mathbf{0.641\pm0.089}$ & $0.582\pm0.003$ & $\mathbf{0.085\pm0.012}$ \\
KPGT MT & $0.622\pm0.012$ & $\mathbf{0.817\pm0.007}$ & $0.550\pm0.040$ & $0.589\pm0.008$ & $\mathbf{0.128\pm0.037}$ \\
KERMT MT & $\mathbf{0.712\pm0.005}$ & $\mathbf{0.822\pm0.002}$ & $\mathbf{0.666\pm0.025}$ & $\mathbf{0.683\pm0.008}$ & $\mathbf{0.096\pm0.014}$ \\

\botrule
\end{tabular*}
\footnotetext{The best performing models, including those with overlapping error bars, are bolded. ST indicates single task models, MT indicates multi-task models, and $n_{\text{train}}$ indicates size of the training data.}
\end{table}

\begin{figure}[h!]
\vspace*{-4.6\baselineskip}
\centering
\includegraphics[width=1.0\textwidth]{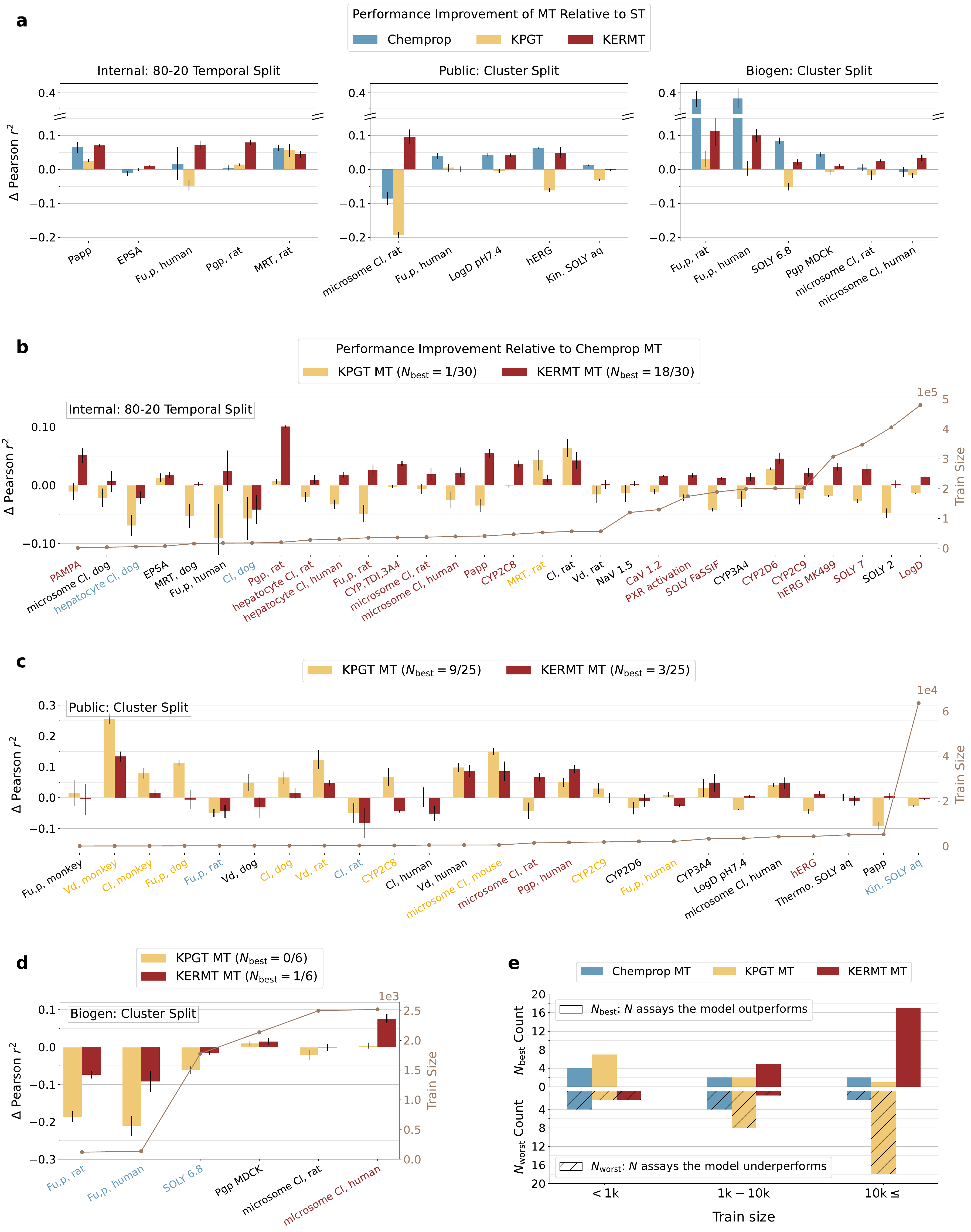}
\caption{\textbf{Comparison of model performance on ADMET assay predictions.} \textbf{a} Performance improvement of multitask (MT) models compared to single-task (ST) models. \textbf{b} - \textbf{d} Performance improvement of pretrained MT models compared to Chemprop MT model. The plots show the results on all available assays for the internal, public, and Biogen datasets, respectively. Brown solid lines represent the number of training data for each assay. A colored assay name (blue: Chemprop, yellow: KPGT, red: KERMT) represents the model which performs statistically better as indicated by the Pearson $r^2$ being higher than the upper error limit of other models. Black labeled assays indicate that two or three models performed statistically the same. $N_{\text{best}}$ represents the number of assays the model statistically outperforms. \textbf{e} Number of assays the model statistically outperforms ($N_{\text{best}}$) or underperforms ($N_{\text{worst}}$) for different ranges of assay training size.}
\label{fig:mt_assay}
\end{figure}

Among the pretrained models utilizing standard single-task finetuning, the KERMT and KPGT models with simple finetuning performed better on our internal data. These models benefit from representing the molecules as a molecular graph, directly capturing connectivity information. As evidenced by MoLFormer's lower performance, pretraining on basic SMILES string representations (without any derived features) does not yield more expressive molecular representations compared to molecular graphs, even when pretraining on a much larger chemical space. This underscores the importance of preserving graphical structure in the prediction of molecular properties. Additionally, we found that the complex finetuning strategy that was recommended in the original paper degrades the performance of the KPGT model \cite{Li2023}. The complex finetuning scheme allows the model to balance between retaining the information learned from pre-training tasks and updating the encoder with downstream tasks. Contrary to this, the simple finetuning method does not use regularization methods, enabling the encoder weights to update more freely and flexibly when finetuning on downstream tasks\toggletext{ (Supplementary Table 7)}. The simple KPGT finetuning model consistently outperforms the complex finetuning model throughout our work.

The single-task variants of KERMT and KPGT models outperform Chemprop single-task on 3/5 and 1/5 assays, respectively. However, these models both only outperform Chemprop multitask on 1/5 assays. It appears that training Chemprop from scratch in a multitask manner yields higher performance compared to finetuning a single-task, pretrained encoder model. When KERMT is trained with a multitask FFN, its performance is the best, or tied for best, on all assays against other methods benchmarked. The performance improvement from utilizing multitask training on each method on the internal dataset is visualized in Fig. \ref{fig:mt_assay}a. The KERMT model improves from multitasking the most overall and does not degrade in performance for any assay. On the other hand, KPGT benefits little from multitasking and even significantly reduces performance for the $F_{u,p}$, human assay. Due to the improvements from multitasking, the multitask variant of KERMT is the best model across all single-task and multitask models benchmarked on these five internal ADMET assays. 

Fig. \ref{fig:mt_assay}b shows the performance improvement by the pretrained multitask models relative to the Chemprop multitask model on our internal temporal split across all assays. KERMT multitask is statistically superior in 18/30 assays, while KPGT is the best in 1/30 assays. Relative to Chemprop multitask, KERMT multitask either improves in or remains the same performance, while KPGT multitask typically decreases in performance for a majority of the assays. KERMT multitask also has a 0.02 and 0.04 larger average Pearson $r^2$ compared to Chemprop and KPGT respectively\toggletext{ (Supplementary Fig. 3a)}. When considering the overall performance across all 30 internal ADMET endpoints in the multitask setting, KERMT multitask remains the best model.

\subsubsection{Benchmarking on public ADMET data}
\label{sssec:public}

We additionally benchmarked each of the models on public ADMET datasets. When benchmarking on our public cluster split, we find that KPGT single task performs the best in 3/5 of the tasks (Table \ref{tab:public}). Contrary to our internal benchmarking, the KERMT multitask model only outperforms the other methods in 1/5 tasks. KPGT is additionally the best when benchmarked against the other multitask methods on all 25 assays in the public cluster split (Fig. \ref{fig:mt_assay}c). KPGT multitask performs the best in 9/25 tasks, while KERMT multitask performs best in 3/25 tasks. Relative to Chemprop multitask, KPGT multitask typically improves in or remains the same performance, while KERMT multitask typically remains the same performance on the public dataset. On the Biogen dataset, both pretrained models performed poorly relative to Chemprop multitask, other than on the human microsome clearance assay where KERMT multitask is better. Due to the small test sizes in the Biogen dataset, there are large error bars yielding low statistical confidence to the results. 

Similar to the internal benchmarking, we find that in both of the public datasets KPGT does not improve from multitasking (Fig. \ref{fig:mt_assay}a). In fact, KPGT decreases in performance for $>50\%$ of the assays across both the public cluster split and the Biogen cluster split. KERMT consistently improves from multitask learning in both public ADMET datasets, though more marginally compared to the internal temporal split. 

\begin{table}[h!]
\caption{Public ADMET data benchmarking of single-task and multitask methods (Pearson $r^2$) on the cluster split}\label{tab:public}
\addtolength{\tabcolsep}{-3pt}
\begin{tabular*}{\textwidth}{@{\extracolsep\fill}lccccc}
\toprule%

 & microsome Cl, rat & $\text{F}_\text{u,p}$, human & LogD & hERG binding & Kinetic SOLY, aq \\
Model & $n_{\text{train}}$=1k & $n_{\text{train}}$=2k & $n_{\text{train}}$=3k & $n_{\text{train}}$=4k & $n_{\text{train}}$=64k \\
 \midrule
MolFormer ST & $0.168\pm0.027$ & $0.593\pm0.013$ & $0.596\pm0.006$ & $0.353\pm0.025$ & $0.602\pm0.002$ \\
Chemprop ST & $0.338\pm0.041$ & $0.618\pm0.017$ & $0.695\pm0.011$ & $0.443\pm0.013$ & $0.676\pm0.005$ \\
KPGT ST\textsubscript{complex} & $0.337\pm0.006$ & $0.641\pm0.007$ & $0.624\pm0.004$ & $0.434\pm0.010$ & $0.619\pm0.002$ \\
KPGT ST & $\mathbf{0.403\pm0.013}$ & $\mathbf{0.662\pm0.009}$ & $0.703\pm0.011$ & $\mathbf{0.524\pm0.015}$ & $\mathbf{0.692\pm0.001}$ \\
KERMT ST & $0.223\pm0.028$ & $0.631\pm0.008$ & $0.702\pm0.006$ & $0.470\pm0.014$ & $0.686\pm0.002$ \\
 \midrule
Chemprop MT & $0.252\pm0.039$ & $\mathbf{0.658\pm0.002}$ & $\mathbf{0.738\pm0.008}$ & $\mathbf{0.506\pm0.013}$ & $\mathbf{0.689\pm0.002}$ \\
KPGT MT & $0.210\pm0.025$ & $\mathbf{0.668\pm0.017}$ & $0.699\pm0.011$ & $0.462\pm0.019$ & $0.662\pm0.007$ \\
KERMT MT & $0.318\pm0.021$ & $0.632\pm0.013$ & $\mathbf{0.742\pm0.012}$ & $\mathbf{0.520\pm0.025}$ & $0.684\pm0.004$ \\

\botrule
\end{tabular*}
\footnotetext{The best performing models, including those with overlapping error bars, are bolded. ST indicates single task models, MT indicates multi-task models, and $n_{\text{train}}$ indicates size of the training data.}
\end{table}

\begin{table}[h!]
\caption{Biogen ADMET data benchmarking of single-task and multitask methods (Pearson $r^2$) on the cluster split}\label{tab:biogen}
\addtolength{\tabcolsep}{-3pt}
\begin{tabular*}{\textwidth}{@{\extracolsep\fill}lcccccc}
\toprule%
 &  &  &  &  & microsome & microsome \\
 & $\text{F}_\text{u,p}$, rat & $\text{F}_\text{u,p}$, human & SOLY 6.8 & P-gp MDCK & Cl, rat & Cl, human \\
Model & $n_{\text{train}}$=120 & $n_{\text{train}}$=135 & $n_{\text{train}}$=2k & $n_{\text{train}}$=2k & $n_{\text{train}}$=2k & $n_{\text{train}}$=3k \\
 \midrule
MolFormer ST & $0.119\pm0.023$ & $0.269\pm0.035$ & $0.181\pm0.010$ & $0.415\pm0.007$ & $0.570\pm0.015$ & $0.510\pm0.004$ \\
Chemprop ST & $0.121\pm0.085$ & $0.190\pm0.012$ & $0.296\pm0.027$ & $0.475\pm0.009$ & $\mathbf{0.596\pm0.014}$ & $0.566\pm0.022$ \\
KPGT ST\textsubscript{complex} & $0.269\pm0.043$ & $0.291\pm0.031$ & $0.337\pm0.003$ & $\mathbf{0.520\pm0.007}$ & $0.548\pm0.013$ & $0.505\pm0.004$ \\
KPGT ST & $0.284\pm0.064$ & $0.358\pm0.062$ & $\mathbf{0.369\pm0.012}$ & $\mathbf{0.538\pm0.012}$ & $\mathbf{0.597\pm0.013}$ & $0.580\pm0.009$ \\
KERMT ST & $0.314\pm0.064$ & $0.381\pm0.018$ & $0.343\pm0.009$ & $\mathbf{0.525\pm0.009}$ & $\mathbf{0.577\pm0.017}$ & $0.600\pm0.019$ \\
 \midrule
Chemprop MT & $\mathbf{0.501\pm0.046}$ & $\mathbf{0.572\pm0.059}$ & $\mathbf{0.380\pm0.012}$ & $\mathbf{0.519\pm0.012}$ & $\mathbf{0.602\pm0.011}$ & $0.558\pm0.019$ \\
KPGT MT & $0.315\pm0.020$ & $0.362\pm0.019$ & $0.318\pm0.013$ & $\mathbf{0.529\pm0.007}$ & $\mathbf{0.580\pm0.020}$ & $0.562\pm0.007$ \\
KERMT MT & $\mathbf{0.427\pm0.038}$ & $\mathbf{0.481\pm0.040}$ & $\mathbf{0.365\pm0.008}$ & $\mathbf{0.535\pm0.009}$ & $\mathbf{0.602\pm0.011}$ & $\mathbf{0.634\pm0.010}$ \\

\botrule
\end{tabular*}
\footnotetext{The best performing models, including those with overlapping error bars, are bolded. ST indicates single task models, MT indicates multi-task models, and $n_{\text{train}}$ indicates size of the training data.}
\end{table}

\subsubsection{Analysis on internal and public ADMET benchmarking}
\label{sssec:analysis}
Our findings on each of the datasets can be explained mostly by the training size, though there is still variability that could be explained by other factors such as the correlation and data overlap between the tasks. Contrary to prior hypotheses on chemical pretrained models, we observe that KERMT performs best on assays of larger data size rather than small data sizes. Fig. \ref{fig:mt_assay}e shows the best and worst multitask models for three data ranges across all three ADMET datasets: (1) small ($<1$k), (2) medium (1k-10k), and (3) large ($>10$k).

\subsubsection{Benchmarking on on-target potency data}
\label{sssec:potency}

\begin{figure}[h!]
\centering
\includegraphics[width=0.82\textwidth, height=0.82\textheight]{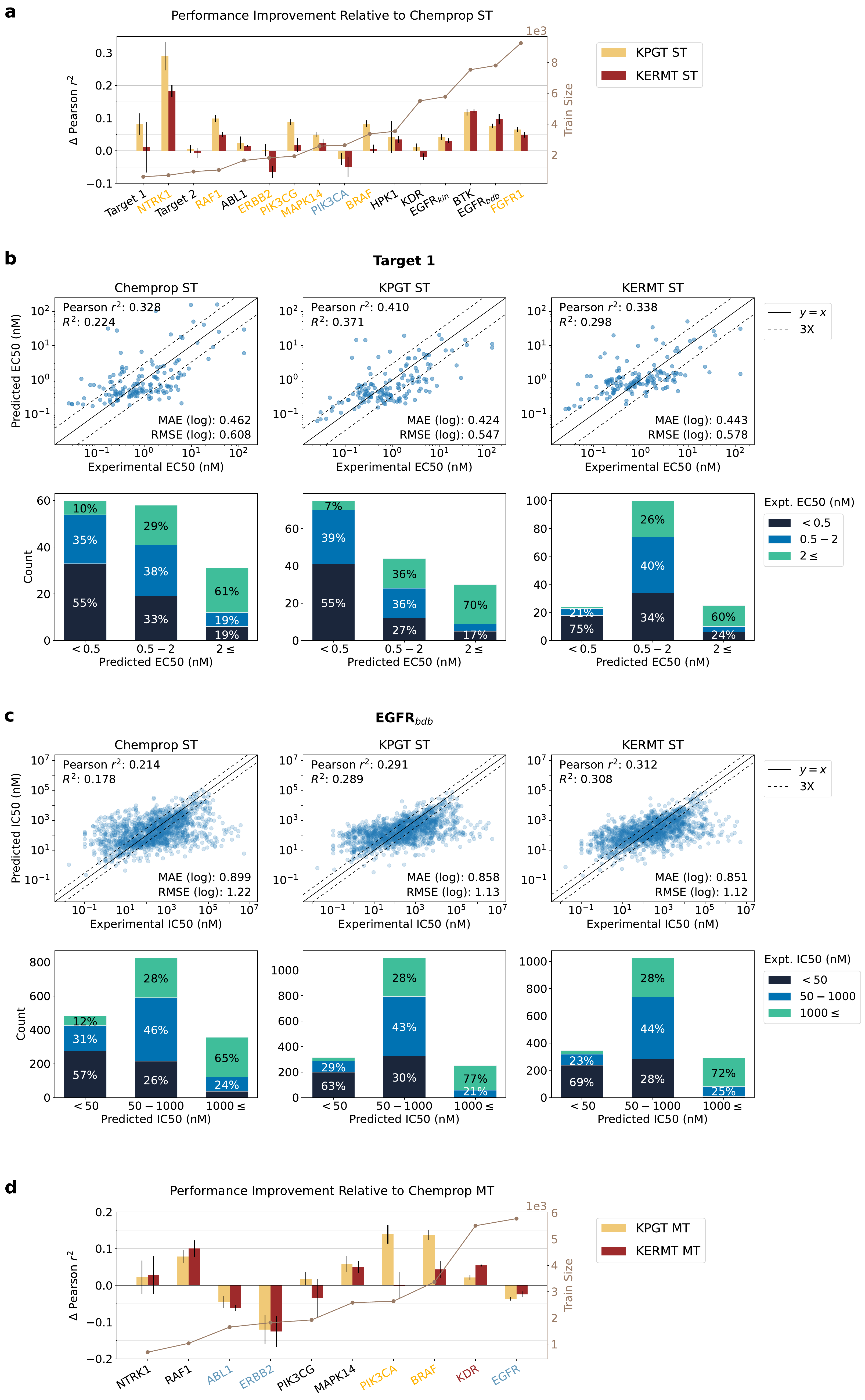}
\caption{\textbf{Model performance for on-target potency prediction.} \textbf{a}, \textbf{d} Performance improvement for pretrained models compared to Chemprop. Brown line indicates number of training data for a target. \textbf{a} Results on single task potency models. EGFR$_{kin}$ is from the kinase multitask dataset and EGFR$_{bdb}$ is from BindingDB. \textbf{d} Results on multitask kinase potency model. \textbf{b} - \textbf{c} Parity plots (top) and categorical bar plots (bottom) for each model. $R^2$, MAE, and RMSE represent coefficient of determination, mean absolute error, and root mean squared error in $\log_{10}(\mathrm{nM})$ units. The categorical bar plot is annotated with the percentage of the predicted bin (x-axis) that belongs to each experimental bin (colored legend). \textbf{b} $EC_{50}$ predictions for Target 1. \textbf{c} $IC_{50}$ predictions for EGFR.}
\label{fig:potency}
\end{figure}

Traditionally, it is thought that chemical pretrained models would be able to improve potency predictions as these are typically in the low-data range. Potency measurements are typically taken for molecules of promising series due to their high cost leading to small datasets relative to our ADMET data. Though more classical models like Random Forest are typically used in low-data scenarios, chemical pretrained models may be more effective in capturing high complexity for these difficult tasks. On-target potency is typically modeled using single-task methods as the mechanisms of action can vary drastically between targets. Thus, we benchmarked each of the single task models on six on-target potency datasets corresponding to different targets. 

We observe relatively similar data scaling trends of KPGT and KERMT utilizing both Pearson ${r^2}$ and categorical enrichment plots to measure performance. The average Pearson ${r^2}$ indicates that KPGT performs better at data sizes $<3K$ and KERMT generally improves as the data size increases. It should be noted that due to the small testing size there is low confidence in the results for some endpoints, as indicated by the large error bars. Parity plots and categorical bar plots for Target 1 and EGFR$_{bdb}$ potency predictions are displayed in Fig. \ref{fig:potency} and can be used to determine the model utility ranking, especially when Pearson ${r^2}$'s are similar. \toggletext{The corresponding plots for the other targets are shown in Supplementary Fig. 4. }The categorical bar plots compare if the model can correctly categorize the compounds as highly potent (low EC50/IC50, black bin), medium potent (medium EC50/IC50, blue bin), or not potent (high EC50/IC50, teal bin). These categorical plots also show evidence that KPGT performs best for small datasets (Target 1) and KERMT performs best for larger datasets (EGFR$_{bdb}$), even when their Pearson ${r^2}$ values are relatively similar. In Fig. \ref{fig:potency}b, KPGT classifies more highly potent molecules correctly leading to lower percentages of misclassified highly potent molecules in the two larger EC50 bins. In Fig. \ref{fig:potency}c, all models classify approximately the same number of highly potent molecules correctly, but KERMT has less misclassifications leading to better enrichment for identifying potent molecules. Additionally, we benchmark the chemical pretrained models against Boltz-2, a pretrained co-folding model \cite{Passaro2025}. The performance of Boltz-2 is significantly worse than the chemical pretrained models and Chemprop in most assays, but it should be noted that this model is not finetuned on labeled data like the other models. 

Though on-target potency is traditionally modeled using single-task methods, we additionally benchmark these models on a kinase on-target potency multitask dataset. There is not a clear best model for this kinase multitask set as Chemprop is the best for 3/10 assays, KPGT is the best for 2/10 assays, and KERMT is the best for 1/10 assays.

\subsection{Studies on KERMT model scaling}
\label{ssec:scaling}

\begin{figure}[h!]
\vspace*{-2.5\baselineskip}
\centering
\includegraphics[width=0.9\textwidth]{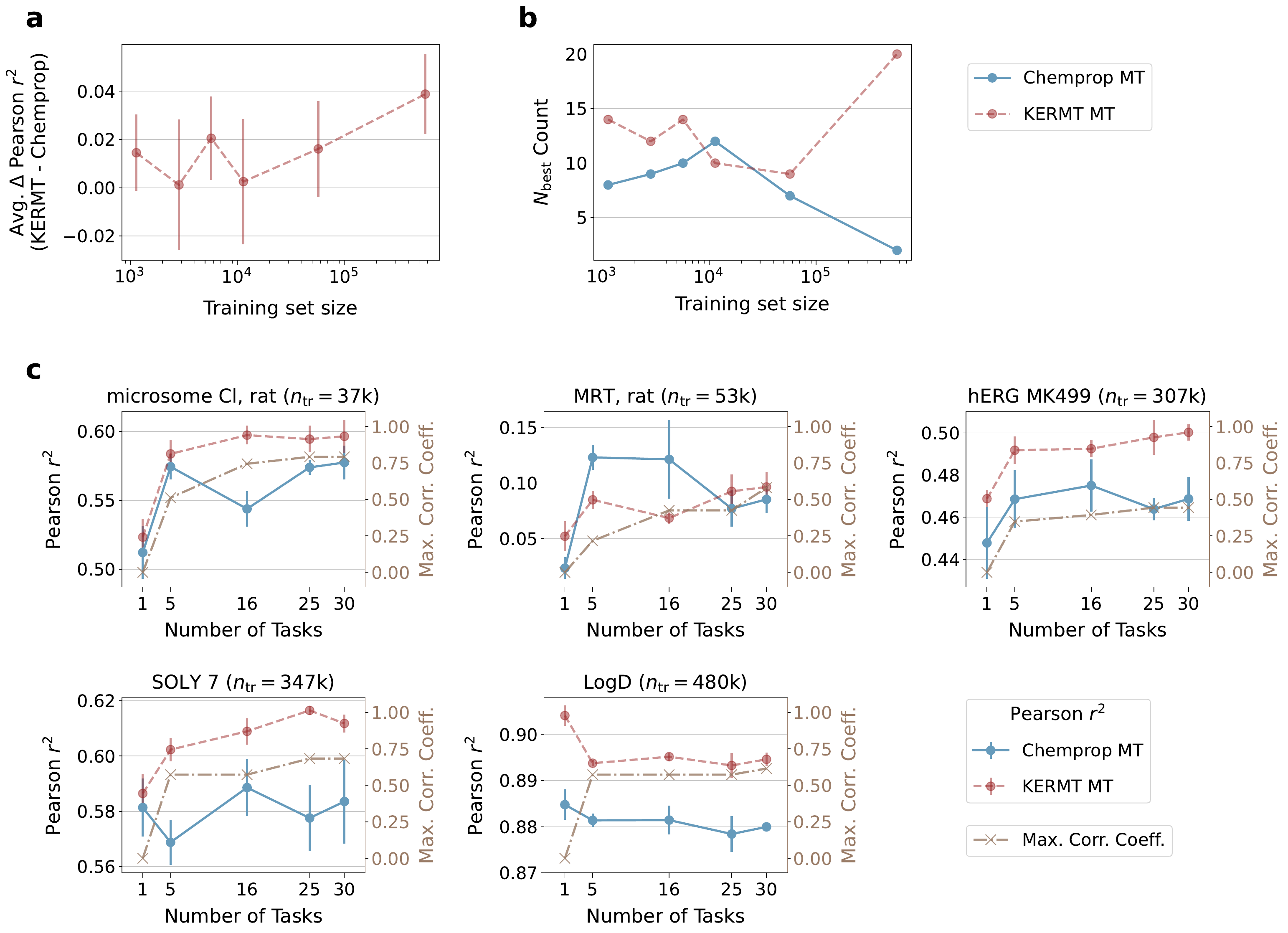}
\caption{\textbf{Scaling of Chemprop multitask and KERMT models with training set size and number of tasks.} All results are evaluated on the internal ADMET dataset with the 80-20 temporal split. \textbf{a} Average performance improvement by the KERMT model relative to the Chemprop MT model for different training set sizes. The average performance improvement is calculated using top 10 assays with the most performance difference between KERMT and Chemprop models. \textbf{b} Number of assays each model statistically outperforms for different training set sizes. \textbf{c} Pearson $r^2$ of each model on selected assays across a different number of tasks in a multitask model. A brown dash-dotted line with cross markers (Max. Corr. Coeff.) represents the maximum value of the absolute Spearman correlation coefficients between the selected assay and all other assays in each multitask dataset.}
\label{fig:scaling}
\end{figure}

We additionally investigated the scaling behavior of KERMT by altering the size of the full chemical space used to finetune the model in the multitask setting. It is hypothesized that chemical pretrained models are better than the Chemprop models that are trained from scratch in the low data regime because there is additional information learned from the unlabeled chemical space used to pretrain the model. Contrary to this and consistent with our observations on scaling based on the individual task data size, we find that KERMT typically performs better at larger full data sizes as shown in Fig. \ref{fig:scaling}a and \ref{fig:scaling}b. As the training size decreases, both the improvement in average Pearson $r^2$ and the number of assays with statistically superior Pearson $r^2$ decrease in KERMT relative to Chemprop. We observe significantly superior performance in KERMT at a full downstream data size ${>60K}$ datapoints. At data sizes ${<60K}$, both models perform relatively similar, though KERMT minimally outperforms. \toggletext{The training size scaling of Chemprop and KERMT models on individual assays can be found in Supplementary Fig. 12.}

Based on the original publication, KERMT performs best when finetuning both the encoder and FFN weights on a downstream labeled dataset \cite{Rong2020}. This allows the general representations that are learned in the self-supervised step to be altered by the labeled data, generating more optimal representations for the downstream task. Because the KERMT model has a larger trainable parameter size ($\sim$51 million) than the Chemprop model ($\sim$5 million), we believe it benefits most at larger data sizes. A model with a large parameter space is more prone to overfitting when trained with smaller datasets. We confirm this when repeating this data scaling study on the KERMT large model which has approximately double the parameters as the KERMT base\toggletext{ (Supplementary Fig. 8-9)}. The KERMT base model outperforms the KERMT large model, and the performance difference becomes more pronounced as the training size decreases. 

Additionally, we investigated how scaling the number of tasks in a multitask model affects the performance with respect to both the KERMT and Chemprop multitask models (Fig. \ref{fig:scaling})c. There is a slight improvement in performance when increasing the number of tasks in KERMT, while Chemprop’s performance typically remains the same or marginally worse. As the number of tasks increases, the data count also increases, likely contributing to the increase in performance of KERMT. Additionally, the improvement in KERMT can be attributed to the increase in the maximum correlation across the task space as more tasks are added to the dataset. This trend is not observed for Chemprop, thus suggesting that KERMT is able to more strongly capture correlations between assays due to the initial knowledge gained from its pretraining task. Both the increased data count and the cross-task correlation are influential to KERMT's improved performance.

\subsection{Chemical space generalization of the KERMT model}
\label{ssec:generalizability}

\begin{figure}[h!]
\vspace*{-2.5\baselineskip}
\centering
\includegraphics[width=1.0\textwidth]{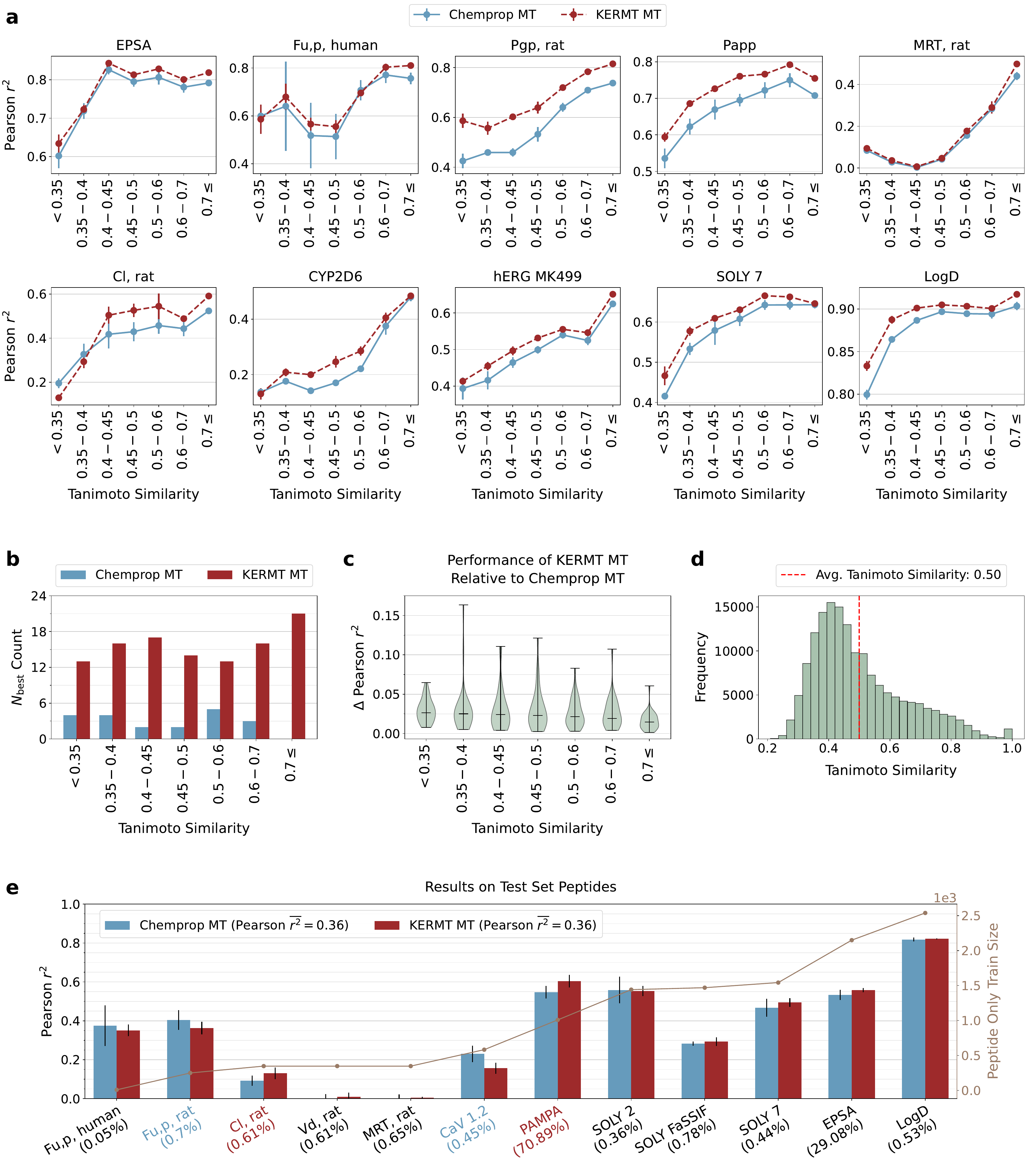}
\caption{\textbf{Chemprop and KERMT multitask model performance across a different range of the Tanimoto similarity scores of the test set compounds and peptides.} All results are evaluated on the internal ADMET dataset with the 80-20 temporal split. \textbf{a} Pearson $r^2$ of each model for selected ADMET assays. \textbf{b} Number of assays each model statistically outperforms for each Tanimoto similarity range. \textbf{c} Violin plot displaying the distributions of the performance improvement by the KERMT model relative to the Chemprop MT model. The central horizontal line in each violin plot represents the median of the distribution. In \textbf{b} and \textbf{c}, the results are evaluated using all 30 assays from the internal dataset. \textbf{d} Histogram of the Tanimoto similarity scores of the test set compounds. \textbf{e} Model performance on the test set peptides. The percentage in the assay label represents the percentage of peptides in the training data for each assay. A brown solid line represents the number of peptide training data for each assay. A colored assay name (blue: Chemprop, red: KERMT)  represents the model which performs statistically better. Pearson $\overline{r^2}$ represents the average performance over all assays.}
\label{fig:perf_vs_sim}
\end{figure}

The information learned during pretraining of molecular property prediction models is thought to allow the model to be more generalizable to less similar chemical space relative to an un-pretrained model like Chemprop. Thus, we compared the performance of KERMT and Chemprop multitask models on subsets of the test set identified by their maximum Tanimoto similarity to the training set (Fig. \ref{fig:perf_vs_sim}a). For each molecule in the test set, we calculated the maximum Tanimoto similarity to the training set molecules based on a 1024-bit Morgan fingerprint. The maximum Tanimoto similarity distribution of the test relative to the training set can be visualized in Fig. \ref{fig:perf_vs_sim}d. As expected, the Pearson $r^2$ of both 
models typically increase as the Tanimoto similarity increases. Additionally, KERMT is consistently better across all Tanimoto bins compared to Chemprop (Fig. \ref{fig:perf_vs_sim}b-c). Though KERMT is not preferentially beneficial to low-similarity compounds, it is the more generalizable model across the entire tested chemical space and yields improvements to all similarity regimes.  

For a subset of the 30 tasks such as solubilities, rat in vivo assays, and $F_{u,p}$, the datasets contain both small molecules and cyclic peptides. The dataset majorly consists of small molecules with molar weight between 200 – 1000 Da whereas it contains relatively small amount of cyclic peptide data with molar weight between 1000 - 2500 Da. We sought to determine if KERMT has strengthened generalization capabilities when predicting properties of peptides, a less represented drug modality in these ADMET datasets. Fig. \ref{fig:perf_vs_sim}e shows the performance on the test set peptides for the tasks with greater than 100 test datapoints using the 80-20 internal split. These results are obtained from the same models shown in Fig. \ref{fig:mt_assay}b, which were trained on full training data containing both small molecules and peptides. Peptides are identified as molecules with six or more peptide bonds. There does not appear to be a significant difference in performance between the two models for cyclic peptide predictions. Currently, the pretrained model has only been trained on small molecules from the ChEMBL \cite{Gaulton2012} and ZINC15 \cite{Sterling2015} repositories \cite{Rong2020}. We explore the impacts of pretraining on chemical space that is more similar to the downstream data in Section \ref{ssec:pretraining}. In the future, it would be interesting to investigate extended pretraining tasks specifically geared towards downstream peptide property prediction.

\subsection{Pretraining on more similar data }
\label{ssec:pretraining}

\begin{figure}[h!]
\centering
\includegraphics[width=\textwidth]{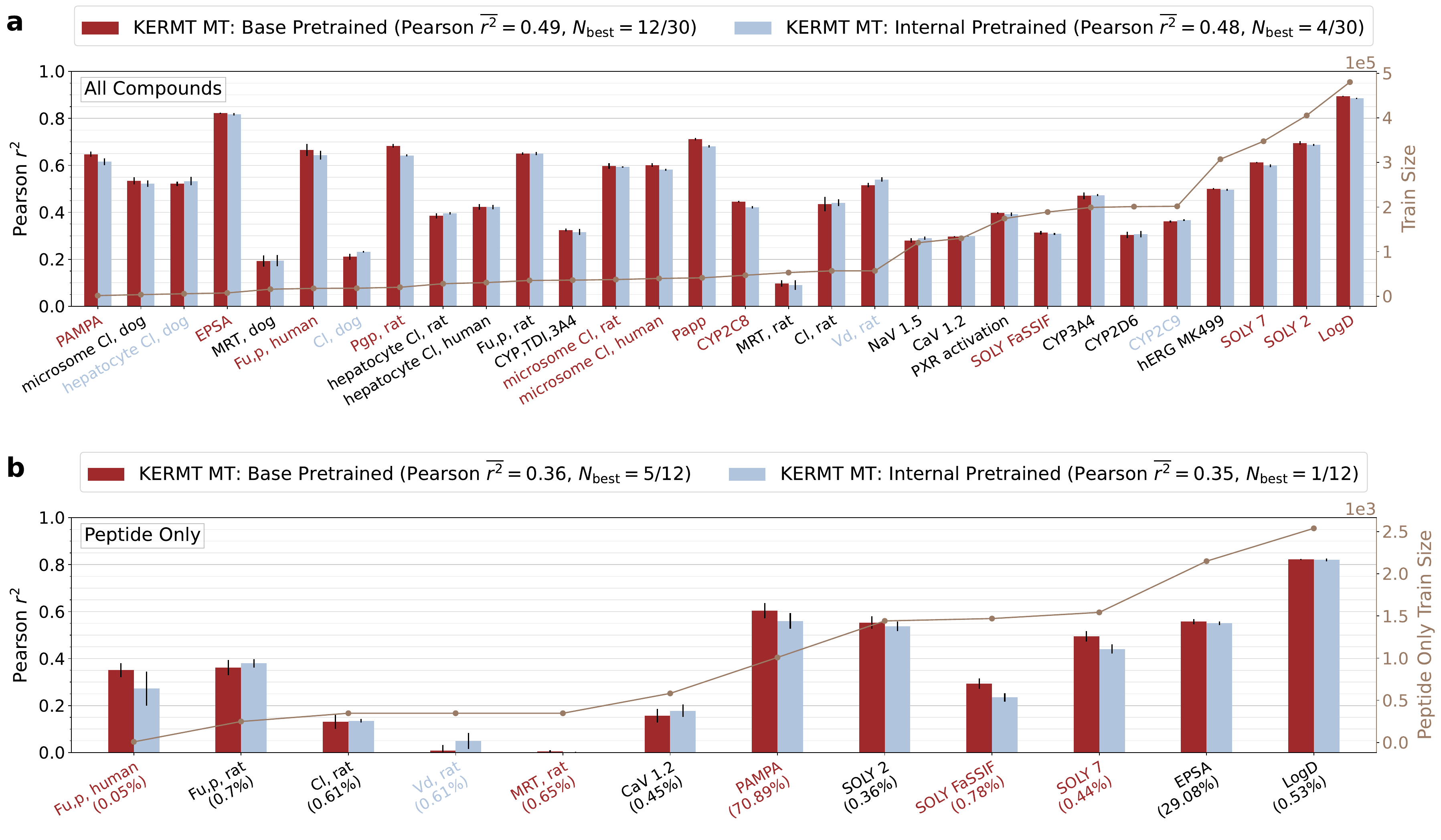}
\caption{\textbf{Comparison of downstream performance after finetuning the KERMT base model (published) and the KERMT model pretrained on our internal compound data.} All results are evaluated on the internal ADMET dataset with the 80-20 temporal split. \textbf{a} Model performance on all test set compounds. \textbf{b} Model performance on the test set peptides. The percentage in the assay label represents the percentage of peptides in the all training data and peptide only training data, respectively, for each assay. In \textbf{a} and \textbf{b}, a brown solid line represents the number of training data for each assay. A colored assay name represents the model which performs statistically better. Pearson $\overline{r^2}$ represents the average performance over all assays, and $N_{\text{best}}$ represents the number of assays the model statistically outperforms.}
\label{fig:grover_pretrained}
\end{figure}

In previous studies on our internal data, we finetuned the pretrained model released from the original publications \cite{Ross2022, Li2023, Rong2020}. Though these models were trained on large corpuses of chemical data from public repositories like ChEMBL \cite{Gaulton2012}, ZINC \cite{Irwin2005}, and PubChem \cite{Kim2023}, these sources do not contain many of the chemical moieties represented in our internal datasets. We pretrained a KERMT model using our full set of $\sim$800k molecules (train and test molecules) from the downstream finetuning tasks to determine whether pretraining on more similar data affects downstream performance. 

It appears that even though the internal pretrained model contained the chemical moieties represented downstream, it did not have a significant effect on downstream performance in terms of average Pearson $r^2$, both for all compounds and the cyclic peptide subset (Fig. \ref{fig:grover_pretrained}). 
The base KERMT model has much less representation of the cyclic peptide drug modality relative to our internal pretraining set, yet the base KERMT model outperforms our internal pretraining on 5/12 of the tasks while our internal pretraining only outperforms the base model on 1/12 tasks. This is surprising as our internal pretrained set has much greater representation of the cyclic peptides relative to the previously published model. We also pretrained a KERMT model using only the training set molecules\toggletext{ (Supplementary Figure 10)}, and it yielded a similar result as the model pretrained on both training and test set molecules. \toggletext{The pretraining losses of two pretrained models (with and without test set compounds) can be found in Supplementary Fig. 11.} 

Removing the test set compounds minimally changes the validation loss curve and the eventual downstream performance. The minimal change in performance for both the cyclic peptide subset and the pretrained model with test molecules indicates that there is more information the model is not extracting in the pretraining stage. Thus, improving the pretraining task to learn the information that is not currently captured is necessary for further improvements. It is also possible that improvements can be made if the model is pretrained with a larger chemical space from both public datasets and internal molecule data. Since pretraining with such large datasets ($>10M$) is a nontrivial task and requires substantial computational resources, it is crucial that we collectively identify the optimal data size and molecule selection scheme for the pretraining dataset. 

\subsection{Model speed-ups}
\label{ssec:speedup}

In general, multitask models improve the computational resources required for training and inference in an industrial setting. At Merck \& Co., Inc. (Rahway, NJ, USA), we have a large labeled chemical space ($>$800k) and over 30 relevant ADMET endpoints. Rather than training 30 individual models, multitasking allows a single model to train for equivalent training time (per molecule, per epoch). This saves compute time, especially when there are multiple individual assays with large data sizes.

Though KERMT in the multitask setting enables these same benefits, finetuning KERMT takes much longer compared to Chemprop for training and inference due to its large parameter size. This makes regularly updating the models with new assay data extremely costly and inefficient. In order to partially alleviate this, we have accelerated the original GROVER implementation\cite{grovercode} and enabled distributed pretraining.

\begin{figure}[h!]
\centering
\includegraphics[width=1.0\textwidth]{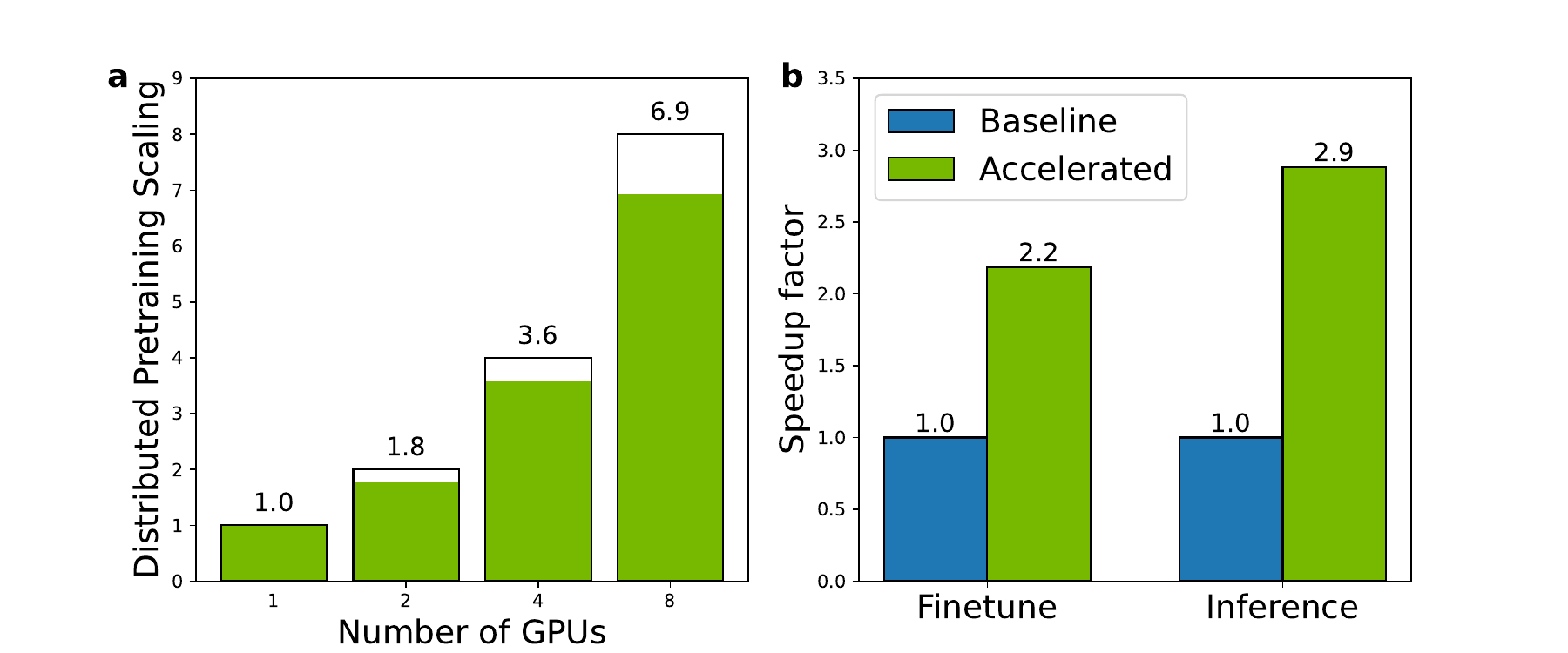}
\caption{\textbf{KERMT model parallelization and acceleration.} \textbf{a} Scaling efficiency of pretraining KERMT model parallelized across multiple GPUs. \textbf{b} Acceleration of KERMT model for finetuning and inference using \texttt{cuik-molmaker} for batched molecular graph generation in dataloader.}
\label{fig:grover_scaling_speedup}
\end{figure}

The pretraining stage of the KERMT model, performed on a large corpus of unlabeled data of 11M molecules, represented a significant computational bottleneck. In order to accelerate pretraining, pretraining was parallelized over multiple GPUs using PyTorch's Distributed Data Parallel (DDP) strategy.\cite{Li2020pytorch} Figure \ref{fig:grover_scaling_speedup} shows the parallelization efficiency of pretraining scaled on multiple GPUs on a single node containing eight NVIDIA A100 GPUs and 16 AMD EPYC 7J13 CPUs. The distributed pretraining implementation parallelizes very efficiently, showing $1.8$X speedup on two GPUs which translates to $90\%$ scaling efficiency. When scaled up to eight GPUs, the scaling efficiency reduces only marginally to $86\%$, showing very efficient use of all eight GPUs during pretraining.

We made use of \texttt{cuik-molmaker}\cite{cuikmolmaker}, a C++/python package that accelerates the featurization of molecular graphs in graph neural networks. This package, capable of producing atom, bond, and molecule-level features for the entire mini-batch at once, was integrated into the dataloader of the KERMT model. For on-the-fly generation of molecular graphs and their corresponding RDKit descriptors, this integration resulted the end-to-end finetuning acceleration by a factor of $2.2$ and inference by $2.9$ on one AMD Ryzen Threadripper CPU core and NVIDIA RTX A6000 GPU. These accelerations were measured for a batch size of 32 and larger speedups are expected for bigger batch sizes. In additional to performance acceleration, on-the-fly generation of molecular graphs and descriptors also reduces CPU memory utilization. For datasets of size comparable to that of the internal datasets at Merck \& Co., Inc., Rahway, NJ, USA, the measured reduction in memory usage is around $34\%$. This enables finetuning and virtual screening on larger datasets than previously possible. Further information about the integration of \texttt{cuik-molmaker} is available in Section \ref{sssec:speedup_method}.

\section{Discussion}\label{sec:discussion}

Here, we propose leveraging the advantages of multitasking in the finetuning of chemical pretrained models. We implemented multitask versions of GROVER and KPGT, and introduce KERMT, a re-implemented and enhanced version of GROVER with acceleration features. Multitasking significantly improves KERMT's performance over its single-task variants, including GROVER, and other pretrained and non-pretrained chemical property prediction models in both single-task and multitask modes. We evaluate the performance of multitask models on public data sets where we generated cluster splits suitable for multitask benchmarking, ensuring no overlap between the test and train sets across all tasks. We also evaluate the performance on a large, temporally split industrial ADMET dataset that is internal to Merck \& Co., Inc. (Rahway, NJ, USA). 

Contrary to prior hypotheses, we find that the performance improvement in KERMT over Chemprop comes at a larger data size (above \textasciitilde10K for individual tasks) making it suitable for large-scale ADMET datasets found in larger drug discovery organizations. The performance improvement in KERMT is observed across all molecule similarity bins tested (Fig. \ref{fig:perf_vs_sim}), showing its increased generalization capabilities. Due to differences in the pretraining tasks, the graph-based pretrained models, KERMT and KPGT, exhibit differing behaviors with respect to both their data scaling and response to multitasking. KPGT's high performance in single-task and small-data settings makes it especially suitable for on-target potency tasks. To enable efficient use in an industrial setting with large ADMET datasets, we accelerate both the pretraining and finetuning algorithm for KERMT and openly release the code.  

KERMT significantly outperforms the other multitask methods, including Chemprop MT and our multi-task implementation of KPGT, in the large data range and more marginally outperforms the other methods in the medium data range. Additionally, regardless of data size KERMT is rarely the worst performing model on a specific assay. On the other hand, KPGT outperforms the other models in the small data range while increasingly performing the worst as the data size increases. The difference likely is due to their varying pretraining tasks. It is possible that the RDKit features learned in KPGT are more beneficial for smaller datasets, where the model relies more on what is learned from a large corpus of unlabeled data. When larger labeled datasets are available, it is sufficient and perhaps more beneficial to learn similar information from a larger labeled dataset as is done with KERMT. To define the degree of "large" where KERMT outperforms non-pretrained, but still multi-tasked, GNNs such as Chemprop, we performed performance tests where we varied dataset size and number of multitask tasks (section \ref{ssec:scaling}) and show that KERMT generally outperforms when training on greater than  five tasks in a multitask fashion and when training with over about 50,000 datapoints. 
 
Comparing multitask versions of KERMT and KPGT, the KERMT model improves the most from multitasking, especially in the internal dataset where the data size is larger and the assays are more correlated, while multitasking KPGT decreases performance for a significant subset of assays in each dataset (see Fig. \ref{fig:mt_assay}a). This is relatively consistent across both internal and public datasets. 

Multitask learning adds generalizability to the model by training on multiple related endpoints at the same time. Oftentimes, this increases the chemical space that the model is trained on as datasets for different assays contain different molecules. Additionally, multitasking can be thought of as a regularization method as the model is forced to learn a shared molecular representation to predict multiple tasks. The general chemical information that KERMT learns in the pretraining stage appears to be especially beneficial to multitask learning. Multitask finetuning is more closely related to the self-supervised pretraining task relative to single-task finetuning as it requires the model to learn general information about each molecule to predict multiple endpoints at the same time. Single-task finetuning requires the model to learn more specific information closely related to the endpoint. Thus, it is possible that the encoder weights are more optimally initialized for the multitask finetuning task, as suggested by its large performance increase.

The disparity in model performances between KERMT and KPGT after multitasking is likely due to the differences in the pretraining tasks. Both models learn about chemical functional groups, but the learning is different. In KPGT, functional groups are learned by predicting a CDF normalized masked descriptor set from RDKit where 85/200 of the descriptors are functional groups.\cite{Li2023} In  KERMT and GROVER, pretraining is performed using RDKit's motif detection algorithm.\cite{Rong2020} However, a difference is apparent in their node-level task where KPGT predicts the identity of masked nodes (atom pairs) while GROVER and KERMT predict the identity of k-hop subgraphs surrounding the node. This likely gives GROVER and KERMT more information on chemical substructures not explicitly identified by RDKit. It is possible that predicting chemical environments directly gives a better starting point for learning general chemical representations for multitask prediction enabling better performance with multitask learning. However, future ablation studies on appropriate pretraining tasks for effective downstream multitasking is likely needed to further elucidate the root cause of this behavior difference. Based on these empirical results, we recommend utilizing KERMT multitask on medium to large data sizes and KPGT single-task on small data sizes for ADMET property prediction.

We find that our performance results for the different models were not preserved across internal and public datasets. A major contributor to this discrepancy is the data size as there is much less data available in the public domain. Additionally, there is much more correlation and overlap between the tasks in our internal multitask set compared to the multitask set we curated from the public domain\toggletext{ (Supplementary Fig. 1)}. This exacerbates the difficulty of developing and ranking models for high-performance in a real-world drug discovery setting. Given the limited multitask ADMET benchmarks currently released and the significant performance improvement observed KERMT in multitask setting, we believe there should be 1) further development of multitasking benchmarks and 2) further benchmarking of chemical pretrained models in the multitask setting.

We plan to further investigate more advanced algorithms, inductive biases, and data curation strategies for the pretraining task that could further benefit downstream performance on additional modalities, such as peptides. Further ablation studies should be conducted to more rigorously determine the correlations between various pretraining tasks and changes in performance from multitasking. Finally, we would like to further investigate how the parameter size and partial freezing for the encoder affects downstream performance at both small and large data sizes.

\section{Methods}\label{sec:method}

\subsection{Datasets}
\label{ssec:datasets}

\subsubsection{Internal datasets}
\label{sssec:internal_datasets}
Each multitask method was trained on internal data collected from Merck \& Co., Inc. (Rahway, NJ, USA) spanning 30 key ADMET endpoints from our drug discovery programs. We chose five important and diverse endpoints from this collection to benchmark single-task methods on: $P_{app}$, EPSA, human $F_{u,p}$, rat P-gp, and rat MRT. Unless otherwise stated, each method was evaluated on an 80\%-20\% train-test split that was split temporally. This split is referenced as the “80-20 split” throughout this study. For the multitask dataset, the temporal cutoff is around April 12th, 2018, where the molecules synthesized after the cutoff are placed in the test set and those synthesized before are placed in the training set. We ensured that no compounds overlap between the training and test sets. 
Similarly, we create the “1-year split” dataset with the time cutoff being January 1st, 2023 where all molecules in the test set were synthesized within one year of this cutoff, and molecules that do not meet this cutoff are placed in the training set. There are 806,556 unique molecules in this dataset, with 21,896 of the molecules being cyclic peptides and the rest being small molecules.\toggletext{ A detailed description of each of the tasks in these data sets and can be found in Supplementary Table 1.} We create single-task dataset splits directly from the multitask split to ensure both models are trained and tested on the same molecules. Additionally, we benchmarked methods on potency data from two undisclosed internal targets which we refer to as “Target 1” and “Target 2”. 

\subsubsection{Public datasets}
\label{sssec:public_datasets}
Each multitask method was trained on 25 ADMET endpoints collected from the literature that are commonly measured in industrial drug discovery programs \cite{Wenzel2019, Iwata2022, Kim2023, Watanabe2018, Falcon-Cano2022, Esposito2020, Braga2015, Aliagas2022, Perryman2020, Meng2022, Vermeire2022}. We chose five diverse end points to benchmark single-task methods on: LogD, hERG binding, rat microsomal clearance, kinetic aqueous solubility, and human $F_{u,p}$. Each method was evaluated on an approximately 80-20 train-test cluster split.\toggletext{ A detailed description of each of the tasks in this data set and their source can be found in Supplementary Table 2.} There are 114,112 unique molecules in this dataset. We reference to this dataset as the “public dataset” throughout this study. 

Additionally, we benchmarked on the previously published dataset from Biogen \cite{Fang2023}. This dataset does not have a published split suitable for multitask evaluation. We also split this dataset on an approximately 80-20 train-test cluster split. \toggletext{ A detailed description of each of the tasks in this data set can be found in Supplementary Table 3.} There are 3,521 unique molecules in this dataset. We refer to this dataset as the “Biogen dataset” throughout this study. 

We also benchmarked methods on public EGFR and BTK $\text{IC}_{50}$ datasets curated from BindingDB (August 2024)\cite{Liu2024}, and literature FGFR1 and HPK1 $\text{IC}_{50}$ datasets\cite{Li2023}. 

To test multitasking for on-target potency dataset, we curate a kinase $\text{IC}_{50}$ dataset taken from a large kinase potency database \cite{Theisen2024}. We chose ten kinase targets to perform multitasking on based on large data size, maximal task correlation, and at least 100 molecules overlapping between the highly correlated assays. \toggletext{ Additional details on the public on-target potency datasets can be found in Supplementary Table 4.} For all public datasets, compounds were clustered into five clusters using K-means based on a PCA-reduced 2048-bit Morgan circular fingerprint (radius 2), and one of the clusters is selected as a test set with the remaining four used as a training set. 

\subsubsection{New public multitask ADMET benchmark dataset}
\label{sssec:public_mt_need}

To our knowledge, there are no large existing public ADMET benchmarks with splits suitable for multitask evaluation. ADMET prediction models are commonly evaluated on datasets from MoleculeNet and, more recently, Polaris \cite{Ash2024, Wognum2024, Wognum2025, Wu2018}. The datasets in the MoleculeNet repository are meant to evaluate single-task models individually with splits created for each task and varying task types (regression and classification). \cite{Wu2018} Additionally, it contains datasets unrelated to ADMET such as quantum mechanical data. The Biogen ADMET set deposited in Polaris does have a random multitask split \cite{Fang2023}, but this is not split by clusters or time. The dataset is also limited in number of compounds (3521 molecules) and coverage of ADMET assays (six endpoints). While time-split is ideal, absent synthesis dates we can try to approximate this using a cluster split to simulate a real-world drug discovery setting where we predict properties of molecules less similar to the training set. Similar limitations are present in the multitask dataset from Wenzel et. al. \cite{Wenzel2019}. We note that \emph{cluster} splits are generally more difficult than \emph{time} splits, and so we are likely underestimating model performance.\cite{Caceres2020} 

We develop a new multitask ADMET benchmark based on public data and are in the process of making this publicly available\toggletext{ on Polaris}. This benchmark is larger ($\sim$120k molecules), has a larger task space (25 endpoints) than previous benchmarks, and is split by clustering. Additionally, we create and release a cluster split for the merged Biogen dataset to enable more rigorous multitask evaluation on this existing dataset.

\subsubsection{Dataset downsampling for model scaling study}
\label{sssec:data_downsampling}
For our scaling experiments we downsampled the datasets both in terms of data size and number of tasks only on the internal 80-20 split. We downsampled the data size with percentages $p \in \{0.2, 0.5, 1, 2, 10\} $. The multitask datasets are inherently sparse with each task (i.e. ADMET endpoint) containing a varying number of datapoints. To ensure that the resulting datasets were less sparse and had enough datapoints for all tasks, we randomly sampled a subset of molecules which had data in n tasks or greater depending on the downsampling percentage\toggletext{ shown in Supplementary Table 5}.

Separately, we downsampled the number of tasks with $n \in \{5, 16, 25\}$ from the original 30 tasks\toggletext{ (Supplementary Table 1)}. The 25-task dataset includes only assays with greater than 5,500 datapoints. There are many highly related endpoints for assays that are performed in multiple host organisms (pharmacokinetic), the same protein family (inhibition), and different solvent pH conditions (solubility). To create the 16-task dataset, we chose a single representative assay from each of these related groupings. To create the 5-task dataset, we chose the assays that represent a diverse range of ADMET properties while including those with large data to ensure the size of the chemical space remains relatively similar to the previous datasets. 
For both of these downsampling studies, we only altered the training set, keeping the test set identical in size to the original 80-20 test set used to benchmark models throughout this study. To compare the effects of downsampling tasks across all scenarios, we only test on the five assays from the $n=5$ downsampling case.

\subsection{Models}
\label{ssec:models}

\subsubsection{Overview of pretrained methods}
\label{sssec:pretraining_methods}

Details of the pretrained models used in this work, KERMT, KPGT, and MoLFormer, are provided below.

KERMT, a graph transformer model based on the GROVER architecture,  uses atom-based and bond-based message passing to learn meaningful chemical representation of molecules\cite{Rong2020}. This model was pretrained using two tasks: (1) node/edge-level classification task which identifies a k-hop local subgraph from  the embedding of the node/edge and (2) Graph-level task multi-label classification task which identifies functional groups present in the molecule from graph embedding. The model was pretrained on 11 million compounds from the ZINC15 \cite{Sterling2015} and ChEMBL \cite{Gaulton2012}. Unless otherwise specified, we evaluated this method using the published 48M parameter base model.

KPGT is a graph transformer that utilizes a molecular-line graph representation. KPGT is pretrained using a knowledge-guided pretraining strategy involving three tasks: (1) Prediction of masked/corrupted nodes and prediction of masked K-node features (2) Prediction of RDKit fingerprints, and (3) Prediction of 200 molecular descriptors generated with RDKit.\cite{Landrum2021} We used the published 100M parameter model checkpoint, pretrained on 2 million molecules from the ChEMBL29 dataset,\cite{Gaulton2017} for finetuning. For KPGT, we employed two finetuning schemes; the simple finetuning model (referred to as "KPGT") was trained by simply allowing the encoder and FFN to update more freely, while the complex finetuning (referred to as "KPGT\textsubscript{complex}") model utilized the complex finetuning strategies presented in the original paper. These complex finetuning strategies include reinitialization of the last $n$ layers of the encoder, layer-wise learning rate decay, and regularization constraining the finetuned weights to be close to  pretrained weights. The simple finetuning scheme was additionally tested as it aligns more closely with the finetuning scheme used for other pretrained models.

For KERMT and KPGT, we used the hyperparameter search package, Optuna \cite{Akiba2019}, to optimize hyperparameters on the five single-task models trained using the internal data from Merck \& Co., Inc. (Rahway, NJ, USA). We found that the optimized hyperparameters were similar among the five single-task models for each pretrained encoder. Therefore, we derive a general set of hyperparameters from this for each pretrained model and report model performance finetuned using these general hyperparameters for all models. \toggletext{ These hyperparameters are reported in Supplementary Table 6-7. All parameters that are not described in these supplementary tables are set to their default value.} \cite{Li2023, Rong2020} Only data excluding the test sets were used for hyperparameter optimization to prevent data leakage.

MoLFormer \cite{Ross2022} is a SMILES-based transformer model utilizing rotational positional embeddings. It uses a traditional masked language modeling objective for pretraining as introduced in BERT \cite{Devlin2019}. We evaluated this method on the public Huggingface model trained on 10\% of the dataset that was used to train the XL version which was used for evaluation in their paper \cite{Wolf2020}. The 10\% model was pretrained on nearly 100 million compounds from ZINC \cite{Irwin2005} and PubChem \cite{Kim2023} datasets. We finetuned the model based on optimized hyperparameters \toggletext{reported in Supplementary Table 8} \cite{Ross2022}. 

\subsubsection{Chemprop overview}
\label{sssec:Chemprop_overview}
Chemprop is a D-MPNN package used for molecular property predictions commonly used in many industrial drug discovery settings \cite{Yang2019, Heid2024}. This architecture is trained from scratch on a labeled dataset and can be trained in either a single task or multitask manner. We use the same hyperparameters developed in a previous study, but standardize epochs to 60 for all single task models and 40 for all multitask models.\cite{Adrian2024}

\subsubsection{Model finetuning and evaluation}
\label{sssec:mt_finetuning_eval}
The pretrained models used in this work were originally designed for single-task finetuning where both the encoder and a new FFN block update their weights to predict one property \cite{Ross2022, Li2023, Rong2020}. Here we adopt the feed-forward neural network output layer to output $n$ values corresponding to the number of properties in the multitask dataset while still updating the encoder during finetuning. 

We measure the performance of each model primarily using the Pearson $r^2$ of the experimentally derived value and the predicted value on the test sets. For model training, the training set is randomly split into 90\% training and 10\% validation sets, where the validation set is used to determine the best training epoch and the best hyperparameters. Predictions are reported using an average of four individual models that are trained with two random training/validation splits and two random weight initializations. Error bars are reported using the standard deviation of the Pearson $r^2$ among four ensembles unless otherwise indicated. For all plots comparing differences in Pearson $r^2$ ($\Delta$ Pearson $r^2$), the error bars correspond to the standard error among four ensembles.

\subsubsection{Model acceleration and parallelization}
\label{sssec:speedup_method}
 The pretraining stage of the KERMT model was parallelized using PyTorch's Distributed Data Parallel scheme \cite{Li2020pytorch}, enabling scaling pretraining to multiple GPUs. 
 In this parallelization regime, the minibatch is divided up and each partition is processed on one GPU. The gradients across all GPUs are gathered at the end of each step and the model weight are updated. As the DDP strategy involves minimal GPU communicated, this parallelized allows us to increase the throughput of pretraining without any significant overhead. The pretraining parallelization shown in \ref{fig:grover_scaling_speedup}(a) was measured on a single node containing eight NVIDIA A100-SXM4-80GB GPUs with a per-GPU batch size of 512.

The molecule featurization involves generating atom, bond, and molecule-level features like 2D descriptors from RDKit. The molecule featurization step was accelerated using \texttt{cuik-molmaker},\cite{cuikmolmaker} a package that enables batched generation of molecular features for graph neural networks. The \texttt{cuik-molmaker} package can be integrated into the dataloader of a model using two methods: (i) Calling \texttt{cuik-molmaker} functionality in the \texttt{\_\_getitems\_\_} method of the \texttt{Dataset} class in PyTorch or (ii) Calling calling the functionality in the collator function and passing it to the \texttt{Dataloader}. We enabled \texttt{cuik-molmaker} in the KERMT model by implementing it in the collator function. The integration also reduced CPU memory usage by generating all molecular graphs and descriptors on-the-fly on an as-needed basis.

\section*{Data availability}
\label{sec:data_availability}
To facilitate experimental reproducibility and enable community validation of multitask models, we have made available the public datasets along with their corresponding train/validation/test splits. These resources are accessible through \href{https://figshare.com/articles/dataset/Datasets_for_Multitask_finetuning_and_acceleration_of_chemical_pretrained_models_for_small_molecule_drug_property_prediction_/30350548}{Figshare} for download and use.

\section*{Code availability}
\label{sec:code_availability}
The KERMT model code base with \texttt{cuik-molmaker} molecule featurization acceleration and parallelization using DDP strategy is available here \href{https://github.com/NVIDIA-Digital-Bio/KERMT}{https://github.com/NVIDIA-Digital-Bio/KERMT}.
The \texttt{cuik-molmaker} code is available at \href{https://github.com/NVIDIA-Digital-Bio/cuik-molmaker}{https://github.com/NVIDIA-Digital-Bio/cuik-molmaker}. Additionally, the \texttt{cuik-molmaker} package can also be installed from \href{https://pypi.nvidia.com/rdkit-2025.03.2_torch-2.6.0/cuik-molmaker}{pypi.nvidia.com}.



\section*{Acknowledgements}

This work was funded by Merck Sharp \& Dohme LLC, a subsidiary of Merck \& Co., Inc., Rahway, NJ, USA, and NVIDIA. We would also like to acknowledge Xin Yu and Kyle Gion for their help facilitating the collaboration.

\bibliography{reference_exported}

\end{document}